\documentclass[10pt,twocolumn,letterpaper]{article}

\usepackage[pagenumbers]{iccv} % To force page numbers, e.g. for an arXiv version

\usepackage{marvosym}
\usepackage{threeparttable}
\usepackage{xspace}
\def\thename{AlphaDrive\xspace}
\usepackage{epsfig}
\usepackage{graphicx}
\usepackage{amsmath,amsfonts}
\usepackage{amssymb}
\usepackage{amsthm}
\usepackage{amsfonts}
\usepackage{bm}
\usepackage{bbm}
\usepackage{array}
\usepackage{multicol}
\usepackage{multirow}
\usepackage{enumitem}
\usepackage{nccmath}
\usepackage{pifont}% http://ctan.org/pkg/pifont
\usepackage{multirow}
\usepackage{array}
\usepackage{adjustbox}
\usepackage{enumitem}
\usepackage{caption}
\usepackage{booktabs}
\usepackage{ragged2e}
\usepackage[ruled,vlined,linesnumbered]{algorithm2e}
\usepackage{dblfloatfix}
\newcommand{\cmark}{\ding{51}}%
\newcommand{\xmark}{\ding{55}}%

\definecolor{HorizonBlue}{HTML}{177cb0}

\newcommand{\boldparagraph}[1]{\vspace{0.2cm}\noindent{\bf #1}}

\newcolumntype{x}[1]{>{\centering\arraybackslash\hspace{0pt}}p{#1}}

\makeatletter
\def\thickhline{%
  \noalign{\ifnum0=`}\fi\hrule \@height \thickarrayrulewidth \futurelet
   \reserved@a\@xthickhline}
\def\@xthickhline{\ifx\reserved@a\thickhline
               \vskip\doublerulesep
               \vskip-\thickarrayrulewidth
             \fi
      \ifnum0=`{\fi}}
\makeatother

\newlength{\thickarrayrulewidth}
\definecolor{darkgreen}{rgb}{0.0, 0.2, 0.13}
\definecolor{darkspringgreen}{rgb}{0.09, 0.45, 0.27}

\makeatletter

\newcommand{\algorithmfootnote}[2][\footnotesize]{%
  \let\old@algocf@finish\@algocf@finish% Store algorithm finish macro
  \def\@algocf@finish{\old@algocf@finish% Update finish macro to insert "footnote"
    \leavevmode\rlap{\begin{minipage}{\linewidth}
    #1#2
    \end{minipage}}%
  }%
}

\DeclareRobustCommand\onedot{\futurelet\@let@token\@onedot}
\def\@onedot{\ifx\@let@token.\else.\null\fi\xspace}

\makeatother

\definecolor{iccvblue}{rgb}{0.21,0.49,0.74}
\usepackage[pagebackref,breaklinks,colorlinks,allcolors=iccvblue]{hyperref}

\title{AlphaDrive: Unleashing the Power of VLMs in Autonomous \\ Driving via Reinforcement Learning and Reasoning}

\author{
Bo Jiang$^{1,\diamond}$ \quad
Shaoyu Chen$^{1,2}$ \quad
Qian Zhang$^{2}$ \quad 
Wenyu Liu$^{1}$ \quad 
Xinggang Wang$^{1,\textrm{\Letter}}$
\vspace{0.3em} \\
\fontsize{10.0pt}{9.84pt}
\textsuperscript{1} Huazhong University of Science and Technology \quad \textsuperscript{2} Horizon Robotics \\
\href{https://github.com/hustvl/AlphaDrive}{ \ttfamily https://github.com/hustvl/AlphaDrive}
}

\begin{document}
\maketitle

\let\thefootnote\relax\footnotetext{$^\diamond$ Intern of Horizon Robotics. $^\textrm{\Letter}$ Corresponding author.}

\begin{abstract}
OpenAI o1 and DeepSeek R1 achieve or even surpass human expert-level performance in complex domains like mathematics and science, with reinforcement learning (RL) and reasoning playing a crucial role. In autonomous driving, recent end-to-end models have greatly improved planning performance but still struggle with long-tailed problems due to limited common sense and reasoning abilities. Some studies integrate vision-language models (VLMs) into autonomous driving, but they typically rely on pre-trained models with simple supervised fine-tuning (SFT) on driving data, without further exploration of training strategies or optimizations specifically tailored for planning. In this paper, we propose \thename, a RL and reasoning framework for VLMs in autonomous driving. \thename introduces four GRPO-based RL rewards tailored for planning and employs a two-stage planning reasoning training strategy that combines SFT with RL. As a result, \thename significantly improves both planning performance and training efficiency compared to using only SFT or without reasoning. Moreover, we are also excited to discover that, following RL training, AlphaDrive exhibits some emergent multimodal planning capabilities, which is critical for improving driving safety and efficiency. To the best of our knowledge, \thename is the first to integrate GRPO-based RL with planning reasoning into autonomous driving. Code will be released to facilitate future research.
\end{abstract}
% \vspace{-5mm}
\section{Introduction}
\label{sec:introduction}

\begin{figure}[ht]
\centering
% \vspace{1mm}
\includegraphics[width=0.98\linewidth]{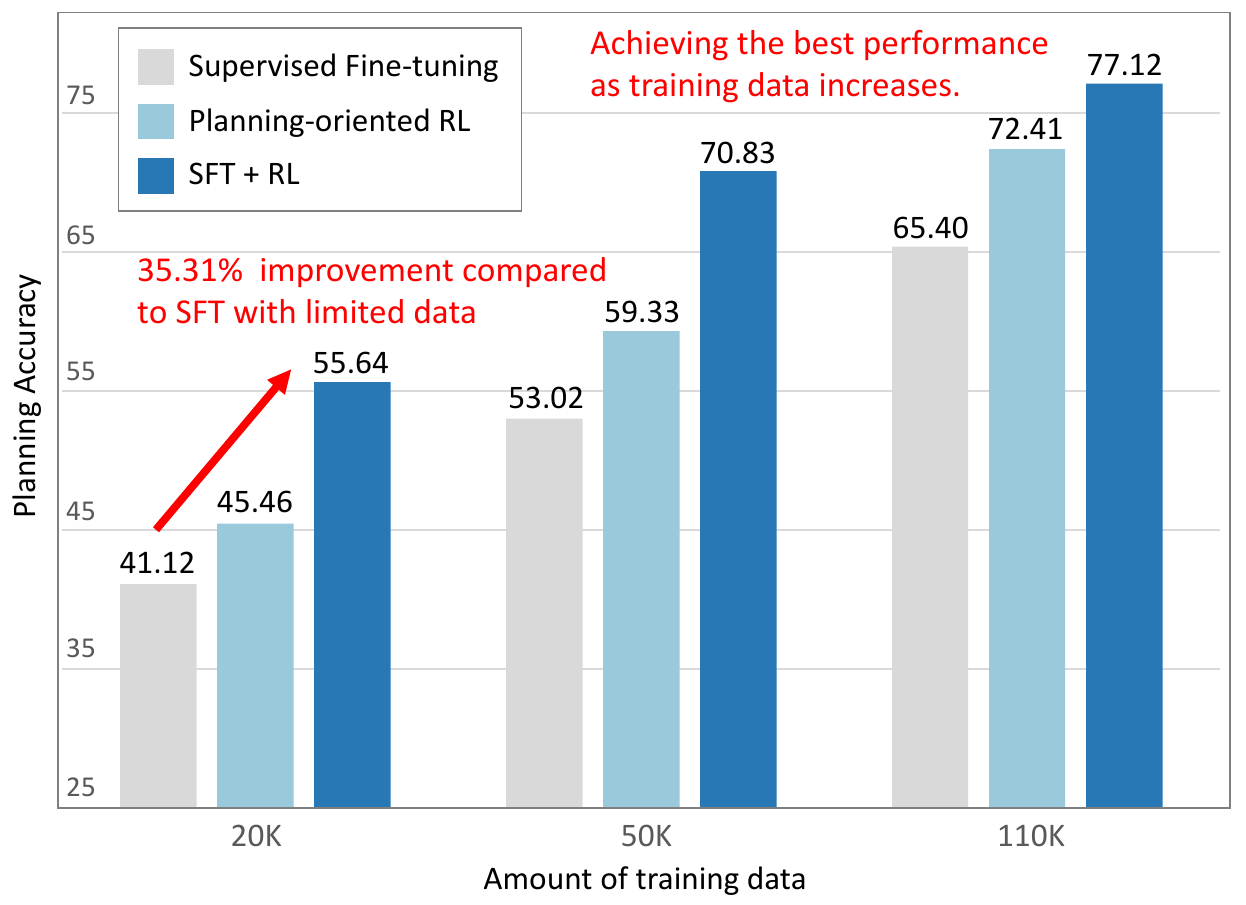} % Reduce the figure size so that it is slightly narrower than the column.
\caption{Our planning-oriented RL and two-stage training strategy significantly boost planning accuracy. With just 20k samples, it outperforms SFT by 35.31\%, showing strong performance even with limited data. As data increase, AlphaDrive consistently leads in planning performance. 
}
\label{fig:teaser}
\end{figure}

Autonomous driving has witnessed rapid advances in recent years, with end-to-end autonomous driving emerging as one of the most representative models~\cite{uniad, vad, chen2024vadv2, transfuser, liao2024diffusiondrive}. They take sensor data as input and leverage learnable neural networks to plan the vehicle's future trajectory. Benefiting from large-scale driving demonstrations, end-to-end models continuously improving their planning capabilities by expanding training data and increasing model parameters.

However, due to their black-box nature and lack of common sense, end-to-end models still face significant challenges when handling complex and long-tail driving scenarios. For instance, consider a situation where the vehicle ahead is carrying traffic cones while driving. An end-to-end model may fail to comprehend the relationship between the leading vehicle and the traffic cones, mistakenly assuming that the road ahead is under construction and thus impassable, leading to an incorrect decision to brake. Therefore, relying solely on end-to-end models to achieve high-level autonomous driving remains challenging.

With the success of GPT~\cite{gpt3}, large language models (LLMs) show remarkable comprehension and reasoning abilities~\cite{touvron2023llama, yang2024qwen2}. Furthermore, their capabilities have evolved from unimodal text understanding to multimodal vision-language processing.~\cite{liu2024llava,chen2024internvl, bai2023qwenvl}. The commonsense and reasoning abilities of VLMs hold great potential to mitigate the limitations of end-to-end models.

Recently, OpenAI o1~\cite{openai-o1}, which incorporates reasoning techniques, achieves performance comparable to or even surpassing that of human experts in fields such as programming. Additionally, DeepSeek R1~\cite{guo2025deepseek-r1}, which leverages reinforcement learning, not only demonstrates \textquotedblleft emergent abilities\textquotedblright and achieves top-tier performance but also requires significantly lower training costs compared to other models. These advances underscore the immense potential of reasoning techniques and RL in the development of large models.

Existing research on applying VLMs to autonomous driving can be broadly categorized into two directions. The first focuses on leveraging VLMs for the understanding of driving scenes~\cite{sima2023drivelm, zhou2024elm}. The second explores the use of VLMs for planning, where some studies treat VLMs as end-to-end systems that process driving images and other inputs to directly predict trajectories~\cite{drivegpt4, driving-with-llms}. However, unlike end-to-end models which are specifically designed for trajectory planning, VLMs operate in a language space and are not inherently suited for precise numerical predictions~\cite{frieder2024mathematical,hendrycks2021measuring}. Consequently, directly employing VLMs for trajectory planning may result in suboptimal performance and even pose safety risks.

Some studies leverage VLMs for high-level planning by formulating the ego vehicle's future actions in natural language, such as \textquotedblleft slow down and turn right\textquotedblright ~\cite{jiang2024senna}. Although this approach circumvents the aforementioned drawbacks, existing works still lack further exploration of training methodologies. Most of them primarily rely on SFT, overlooking the impact of different training strategies on planning performance and the associated training costs.

In this paper, we explore the following question: How can RL and reasoning — which achieves remarkable success in general large models — be applied to autonomous driving, particularly in planning, to enhance the performance of VLMs in autonomous driving while reducing training costs?

Through preliminary experiments, we find that directly applying existing RL and reasoning techniques to planning results in suboptimal performance. We attribute this to three main factors. First, the reward design in RL for general tasks is not well-suited for planning. For example, in visual object counting, the reward can be simply determined based on whether the model predicts the correct answer. However, in autonomous driving, while high-level planning can be formulated as a multi-class classification problem, the varying significance of different driving behaviors makes it inappropriate to assign equal weights to all actions.

Second, unlike mathematical or counting, the solution of planning are usually not unique. For instance, on an open, straight road, one may choose to maintain a constant speed or accelerate, both of which are valid decisions. Therefore, rigidly assessing whether the model's planning output exactly matches the ground truth in the training data may not be the optimal approach.

Finally, while domains such as mathematics have abundant reasoning data, including textbooks and solution manuals that can be easily utilized, autonomous driving lacks readily available datasets that capture the reasoning process. Collecting such data is highly costly and requires extensive manual annotation. As a result, directly applying existing reasoning techniques to planning remains challenging.

To address the aforementioned challenges, this paper introduces \thename, a VLM-based reinforcement learning and reasoning framework specifically designed for autonomous driving planning. In particular, \thename employs a RL strategy based on Group Relative Policy Optimization (GRPO)~\cite{shao2024grpo}. Compared to Proximal Policy Optimization (PPO)~\cite{schulman2017ppo} and Direct Preference Optimization (DPO)~\cite{rafailov2023dpo}, GRPO exhibits better training stability and performance. Furthermore, the group relative optimization strategy in GRPO is well-suited for planning, as planning often involves multiple valid solutions, making relative optimization across multiple solutions a natural fit. Our experiments show that \thename exhibits some emergent multimodal planning capabilities, which we think can be attributed to the use of GRPO.

\thename introduces four GRPO rewards tailored for planning. The first is the planning accuracy reward, which evaluates the consistency between the model's planning actions and the ground truth actions. The second is the action-weighted reward, which assigns different weights to various actions based on their importance to safety. For instance, actions such as braking and steering are critical for safety, so weighting them accordingly helps the model achieve better performance in planning key actions. The third is the planning diversity reward, which encourages the model to generate multiple diverse solutions. This prevents mode collapse and enhances overall planning performance. The last one is the planning format reward, where we define a specific output format and encourage the model to follow it. This ensures more structured outputs and contributes to more stable training.

In addition to RL, we propose a planning reasoning technique. Our approach employs a two-stage training strategy based on knowledge distillation, integrating SFT and RL. In the first stage, we leverage a large model, such as GPT-4o, to generate a small yet high-quality dataset containing planning reasoning processes derived from real driving actions. This dataset is then used to fine-tune our model via SFT, effectively distilling knowledge from the large model. In the second stage, we further refine the model using RL. Introducing the SFT stage as a warm-up step effectively mitigates hallucinations and instability commonly observed in the early stages of reinforcement learning, while also enhancing planning performance.

Our contributions are summarized as follows:
\begin{itemize}
    \item We propose \thename, a VLM tailored for high-level planning in autonomous driving. To the best of our knowledge, \thename is the first to integrate GRPO-based RL with planning reasoning to autonomous driving, significantly boosting both performance and training efficiency.
    \item  \thename introduces four GRPO rewards for planning: planning accuracy reward, action-weighted reward, planning diversity reward, and planning format reward. These optimized rewards make GRPO more suitable for autonomous driving.
    \item We propose a two-stage reasoning training strategy based on knowledge distillation, integrating SFT and RL. Our approach achieves better planning performance compared to training with RL alone or without reasoning. 
    \item Experiments on a large-scale driving dataset validate the superiority of AlphaDrive. Compared to the SFT-trained model, AlphaDrive significantly improves the planning accuracy by 25.52\% and, with only 20\% of the training data, outperforms the SFT-trained model by 35.31\%. We are also excited to discover that, following RL training, AlphaDrive exhibits some emergent multimodal planning capabilities, which is promising for improving driving safety and efficiency.
\end{itemize}

\begin{figure*}[ht]
\centering
% \vspace{1mm}
\includegraphics[width=0.98\textwidth]{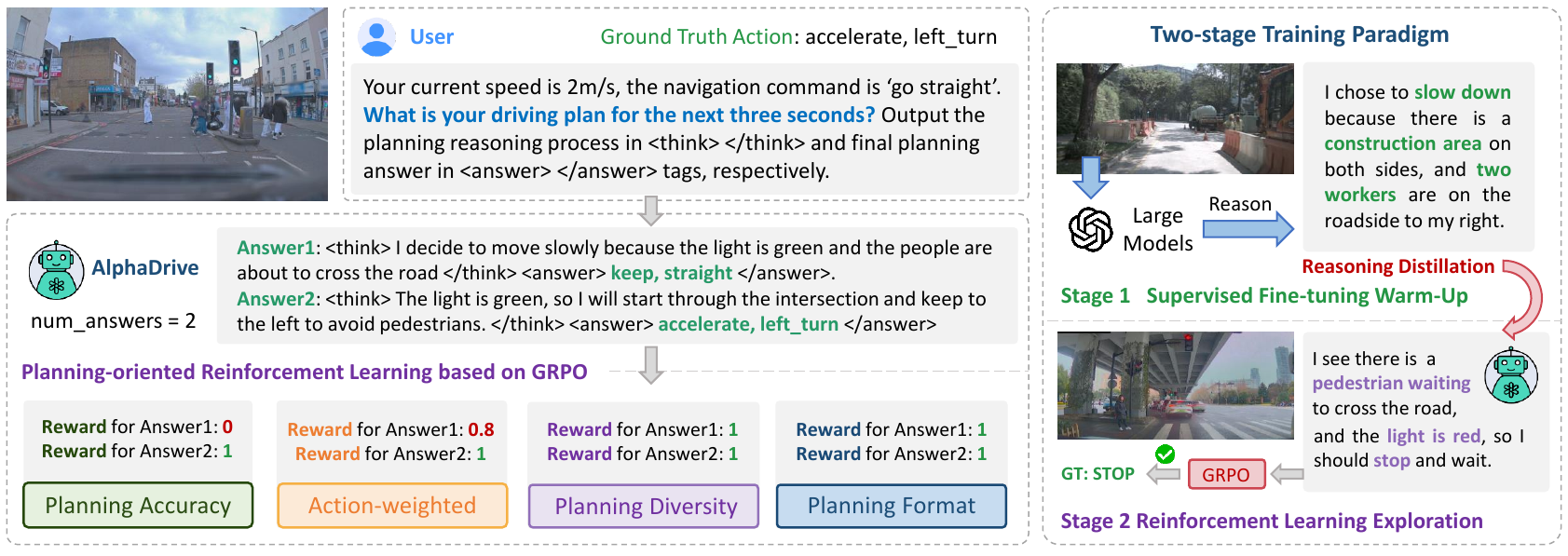} % Reduce the figure size so that it is slightly narrower than the column.
% \vspace{-1mm}
\caption{\textbf{Overall training framework of \thename.}  AlphaDrive is trained using GRPO-based RL, and we design four planning rewards to help the model understand and learn planning. Besides, we propose a two-stage training paradigm, the first stage uses SFT to distill the planning reasoning process from a large model and serves as a warm-up, while the second stage employs RL to explore planning.}
\label{fig:arch}
\end{figure*}
\vspace{2mm}
\section{Related Work}

% \vspace{-1mm}
\boldparagraph{Vision Language Models.} 
Since the release of GPT~\cite{gpt3, achiam2023gpt4}, the capabilities of large models have gradually expanded from single modality to multi-modalities. Large vision language models~\cite{touvron2023llama, achiam2023gpt4, liu2024llava, bai2023qwenvl} now demonstrate superior abilities in visual understanding and reasoning. Early works attempt to integrate visual models with large language models (LLMs), Flamingo~\cite{alayrac2022flamingo} uses a visual encoder to process visual signals and adds attention layers in the LLM decoder to interact with the visual features. BLIP~\cite{li2022blip, li2023blip2} introduces the Q-Former architecture and cross-modal contrastive learning tasks to bridge the vision encoder with LLMs. LLaVA~\cite{liu2024llava, liu2024llava1.5} propose using vanilla MLP as the connector between the visual encoder and LLMs, which achieves impressive visual understanding capabilities with relatively limited data. The QwenVL~\cite{bai2023qwenvl, wang2024qwen2vl} series continuously improve the visual module, offering  better support for high-resolution and dynamic resolution images, while also demonstrate excellent performance in multilingual tasks and spatial perception.

\boldparagraph{Reinforcement Learning and Reasoning.}
Autoregressive learning~\cite{vaswani2017attention} is currently the mainstream pre-training strategy for LLMs. Besides, RL and reasoning techniques further enhance the capabilities of large models~\cite{schulman2017ppo, rafailov2023dpo, ouyang2022rlhf, shao2024grpo, wei2022cot}. For instance, GPT~\cite{achiam2023gpt4} employs RL with Human Feedback (RLHF)~\cite{ouyang2022rlhf}, which incorporates human feedback into the training process. By integrating human intentions and behavioral preferences, RLHF enables LLMs to generate outputs that align more closely with human habits and preferences. Direct Preference Optimization (DPO)~\cite{rafailov2023dpo} enhances the model's performance by directly optimizing preference feedback. Building on this, Group Relative Policy Optimization (GRPO)~\cite{shao2024grpo} introduces a strategy of group relative optimization, which considers the relative superiority or inferiority between multiple output groups, further improving the stability and effectiveness of the training process.

The recent DeepSeek R1~\cite{guo2025deepseek-r1} experiences an \textquotedblleft Aha Moment\textquotedblright during training based on GRPO, where, without any explicit guidance, the model autonomously allocates more thinking to the problem and re-evaluates its initial approach. This highlights the potential of RL in enabling large models to evolve from mere imitation to emergent intelligence. In our experients, we are also excited to discover that, after GRPO-based RL training, AlphaDrive demonstrates some emergent multimodal planning capabilities, enabling it to generate multiple reasonable driving plans. We believe it has great potential to improve driving safety and efficiency.

In terms of reasoning, Chain-of-thought~\cite{wei2022cot} has demonstrated great performance in solving complex problems by breaking them down and reasoning step by step. OpenAI o1~\cite{openai-o1}, which is based on Chain-of-thought, introduces inference-time scaling. By increasing the computational cost during inference and combining search strategies such as Monte Carlo Tree Search (MCTS)~\cite{swiechowski2023mcts} and Beam Search~\cite{xie2023beam}, significant improvements have been achieved in areas such as science and programming that require complex reasoning. This also shows that, beyond scaling model parameters and training data, scaling the inference-time computation is also a promising direction for exploration.

\boldparagraph{Autonomous Driving Planning.} Planning is the ultimate task of autonomous driving. The earliest planning algorithms are rule-based~\cite{paden2016survey, thrun2006stanley}, which have significant limitations in terms of generalizability and efficiency. Recently, end-to-end models~\cite{uniad, vad, chen2024vadv2, transfuser, liao2024diffusiondrive, gao2025rad} has gained popularity, where a unified neural network is used to directly output planning trajectories or control signals from sensor data. By leveraging large-scale driving demonstrations, end-to-end models are trained in a data-driven manner, achieving impressive planning performance. However, since end-to-end models are black-box models that lack common-sense and reasoning capabilities, they still struggle to address the long-tailed problems in autonomous driving.

\boldparagraph{VLMs and Autonomous Driving.}
The common-sense and reasoning abilities of large models can effectively compensate for the limitations of end-to-end models in autonomous driving. In the field of robotics, Vision-Language-Action (VLA) models~\cite{kim2024openvla, brohan2023rt2, wen2024diffusion-vla} have made significant progress in understanding language instructions and executing complex actions. A common approach is to use VLMs as the planning module to generate planning instructions, which are then translated into control signals through an action model. There have also been some works based on large models in the field of autonomous driving. DriveGPT4~\cite{drivegpt4} utilizes a VLM that takes front-view videos as input, and the model directly predicts control signals. ELM~\cite{zhou2024elm} leverages large-scale, cross-domain video training for VLMs, showing that using data from various domains can effectively enhance the performance of VLMs in driving-related tasks. OmniDrive~\cite{wang2024omnidrive} proposes the use of sparse 3D tokens to represent driving scenes, which are then input into VLMs for scene understanding and planning.

% \vspace{-2mm}

In addition to the above works that directly apply VLMs to driving, DriveVLM~\cite{tian2024drivevlm} combines VLMs with end-to-end models for the first time, where VLMs predict low-frequency trajectories and an end-to-end model generates high-frequency trajectories. Senna~\cite{jiang2024senna} proposes a framework where VLMs handle high-level planning, while end-to-end models are responsible for low-level trajectory prediction. Additionally, several datasets and benchmarks have been proposed~\cite{sima2023drivelm, qian2024nuscenes-qa, sima2023drivelm, wu2023nuprompt}, which promote the application of VLMs in autonomous driving. However, most of the current works on VLMs in the field of autonomous driving involves directly using pre-trained models and then utilizing SFT on driving data, which lacks in-depth exploration on training strategies specifically designed for planning. Further effort is needed to adapt the impressive RL and reasoning techniques from general tasks to autonomous driving.

\vspace{-2mm}
\section{\thename}
\label{sec:method}

\subsection{Overview}

AlphaDrive is a VLM designed for autonomous driving planning. Unlike previous approaches that rely solely on SFT, we explore the incorporation of RL and reasoning techniques to better align with the unique characteristics of driving planning: (1) the varying importance of different driving behaviors; (2) the existence of multiple feasible solutions; and (3) the scarcity of readily available reasoning data for planning decisions.

We propose four GRPO-based RL rewards tailored for planning, along with a two-stage planning-reasoning training strategy that integrates SFT with RL. Our experiments demonstrate that, compared to using SFT alone or training without reasoning, AlphaDrive achieves significant improvements in both planning performance and training efficiency. In the following sections, we will detail the design of each component.

\subsection{Planning-oriented Reinforcement Learning}
\subsubsection{Reinforcement Learning Algorithm}

Current commonly used RL algorithms include PPO~\cite{schulman2017ppo}, DPO~\cite{rafailov2023dpo}, and GRPO~\cite{shao2024grpo}. Given a query $q$, GRPO samples a group of outputs $\{o_1, o_2, \cdots, o_G\}$ from the old policy $\pi_{\theta_{old}}$ and optimizes the new policy $\pi_{\theta}$ by maximizing:

\begin{align}
    \mathcal{J}_\text{GRPO}(\theta) &= \mathbb{E}_{q, \{o_i\} \sim \pi_{\theta_{old}}} \left[ \frac{1}{G} \sum_{i=1}^G \mathcal{L}_i - \beta \mathbb{D}_{KL}(\pi_{\theta} || \pi_{ref}) \right], \\
    \mathcal{L}_i &= \min \left( w_i A_i, \text{clip}(w_i, 1 - \epsilon, 1 + \epsilon) A_i \right),
\end{align}

\noindent where $w_i = \frac{\pi_\theta(o_i |q)}{\pi_{\theta_{old}}(o_i |q)}$, $\epsilon$ and $\beta$ are hyper-parameters, and the advantage $A_i$ is computed using the normalized reward within the group. 

We ultimately choose GRPO as the RL algorithm for AlphaDrive for two key reasons: (1) DeepSeek R1~\cite{guo2025deepseek-r1} has demonstrated the effectiveness of GRPO in general domains. Compared to other algorithms, GRPO provides higher training stability and efficiency; (2) Moreover, the group relative optimization strategy introduced by GRPO is particularly well-suited for planning, as planning often involves multiple valid solutions, making relative optimization across multiple solutions is a natural fit. Experimental results further confirm that models trained with GRPO exhibit strong planning capabilities.

\subsubsection{Planning Reward Modeling}
\boldparagraph{Planning Accuracy Reward.}
In fields such as mathematics or programming, the reward in GRPO can be intuitively determined based on whether the final answer is correct. However, planning is more complex, as it involves both lateral (direction) and longitudinal (speed) components. Furthermore, the set of possible actions is constrained. As a result, we use the F1-Score to evaluate the accuracy of both lateral and longitudinal decisions separately, and assign rewards accordingly.

Initially, we evaluate accuracy by checking whether the model's prediction exactly matches the ground truth. However, due to imperfect format in the model’s early training phase, such as discrepancies in case sensitivity or the presence of extraneous outputs, this approach results in poor stability during the early stages of training. We then attempt to extract all the words from the prediction and check whether the ground truth is included among the words. This introduces a new issue where the model sometimes learns shortcut solutions, such as outputting all possible actions, which causes mode collapse. Ultimately, we adopt the F1-score for evaluation, as it not only prevents the model from learning shortcut solutions (where outputting all decisions could result in high recall but low accuracy) but also improves the stability during the early training phase.

\begin{algorithm}[h]
\SetAlgoLined
\DontPrintSemicolon
\SetNoFillComment
\footnotesize
\KwIn{Planning answers $\mathcal{A}$, Ground Truth action $e$}
\KwOut{Planning Reward $\mathcal{R}$}
Initialization: Planning Reward $\mathcal{R} \leftarrow \emptyset$, Action Weights $\mathcal{W}$\;
Speed Action Set $\mathcal{S}$, Path Action Set $\mathcal{P}$, Answer Format $\mathcal{F}$\;
\# Pytorch-like Code\;
ans\_counter = Counter()\;
\For{ans in $\mathcal{A}$}{
    action\_ans = re.search(r\textquotedblleft $\mathcal{F}$\textquotedblright, ans).group(1).strip()\;
    ans\_counter.update(action\_ans)\;
    speed\_ans = \textbf{extract\_ans}(action\_ans, $\mathcal{S}$)\;
    path\_ans = \textbf{extract\_ans}(action\_ans, $\mathcal{P}$)\;
    \# Calculate Planning Accuracy Reward\;
    speed\_acc\_R = \textbf{cal\_f1\_score}(speed\_ans, $e$)\;
    path\_acc\_R = \textbf{cal\_f1\_score}(path\_ans, $e$)\;
    \# Calculate Action-Weighted Reward\;
    speed\_weighted\_R = $\mathcal{W}$\text{[speed\_ans]}\;
    path\_weighted\_R = $\mathcal{W}$\text{[path\_ans]}\;
    \# Calculate Planning Diversity Reward\;
    plan\_div\_R = 0\;
    \If{sum(ans\_counter.values()) != 0}{
        plan\_div\_R = ans\_counter\text{[action\_ans]} / sum(ans\_counter.values())\;
    }
    \# Up to 20\% reduction in diversity reward\;
    plan\_div\_R = 1 - min(0.2, plan\_div\_R)\;
    \# Calculate Planning Format Reward\;
    format\_R = \textbf{check\_format}(\textit{ans}, $\mathcal{F}$)\;
    \# Final Planning Quality Reward\;
    speed\_R = speed\_acc\_R \text{*} speed\_weighted\_R \text{*} plan\_div\_R\;
    path\_R = path\_acc\_R \text{*} path\_weighted\_R \text{*} plan\_div\_R\;
    $\mathcal{R}$.append(\text{[speed\_R, path\_R, format\_R]})\;
}
Return: $\mathcal{R}$
\caption{Planning Reward Modeling.}
\algorithmfootnote{\textbf{extrat\_ans} will extract substrings that match the specified pattern from the given string. \textbf{cal\_f1\_score} will calculate F1 score given the predictions and ground truth. \textbf{check\_format} will check whether the given string matches the provided pattern based on regular expression matching.}
\label{algo:reward}
\end{algorithm}

\boldparagraph{Action-Weighted Reward.}
As mentioned above, the importance of different behaviors in planning varies. For instance, decelerating and stopping are more critical for safety than maintaining speed. Therefore, we assign different importance weights to various actions, incorporating them as weighted components in the final reward.

\begin{table*}
\small
\centering
% \resizebox{0.99\textwidth}{!}{
\setlength\tabcolsep{5pt}
\begin{tabular}{l|c|c|ccc|cccc|ccc}
\toprule
\multirow{2}{*}{Method} & \multirow{2}{*}{Size} & \multirow{2}{*}{Acc. $(\%)$} & \multicolumn{3}{c|}{Path $($F1$)$ $\uparrow$} & \multicolumn{4}{c|}{Speed $($F1$)$ $\uparrow$} & \multirow{2}{*}{BLEU-4} & \multirow{2}{*}{CIDEr} & \multirow{2}{*}{METEOR}\\
& & & straight & left & right & keep & acc. & dec. & stop & & \\
\midrule
InternVL2~\cite{chen2024intern2vl} & 2B & 7.23 & 33.34 & 9.49 & 4.57 & 48.75 & 2.87 & 8.12 & 13.87 & 8.04 & 5.40 & 23.63 \\
Qwen2VL~\cite{wang2024qwen2vl} & 2B & 13.69 & 34.05 & 21.46 & 13.46 & 52.34 & 11.14 & 13.73 & 17.03 & 16.41 & 10.85 & 27.66 \\
Llama3.2-V~\cite{dubey2024llama3} & 11B & 11.61 & 32.56 & 27.78 & 13.67 & 42.71 & 12.77 & 20.81 & 18.04 & 23.23 & 15.87 & 26.30 \\
Qwen2VL~\cite{wang2024qwen2vl} & 7B & 19.28 & 45.92 & 33.09 & 19.20 & 54.13 & 12.86 & 27.01 & 23.48 & 30.30 & 16.16 & 33.36 \\
\midrule
InternVL2${\text{\textdagger}}$~\cite{chen2024intern2vl} & 2B & 51.07 & 76.13 & 85.16 & 64.60 & 74.77 & 21.88 & 47.66 & 15.81 & 27.89 & 19.73 & 28.26 \\
Qwen2VL${\text{\textdagger}}$~\cite{wang2024qwen2vl} & 2B & 55.84 & 82.68 & 80.31 & 70.04 & 75.97 & 34.92 & 55.55 & 72.64 & 24.46 & 23.14 & 34.26 \\
Llama3.2-V${\text{\textdagger}}$~\cite{dubey2024llama3} & 11B & 58.21 & 85.58 & 84.64 & 79.12 & 74.79 & 35.56 & \underline{58.99} & \underline{76.20} & 32.05 & 21.25 & 37.70 \\
Qwen2VL${\text{\textdagger}}$~\cite{wang2024qwen2vl} & 7B & \underline{61.44} & \underline{86.45} & \underline{85.84} & \underline{87.75} & \underline{84.53} & \underline{43.81} & 56.30 & 73.80 & \underline{41.09} & \underline{30.65} & \underline{47.47} \\
AlphaDrive & 2B & \textbf{77.12} & \textbf{96.62} & \textbf{89.83} & \textbf{93.25} & \textbf{86.80} & \textbf{56.33} & \textbf{71.40} & \textbf{86.63} & \textbf{43.54} & \textbf{38.97} & \textbf{55.23} \\
\bottomrule
\end{tabular}
% }
% \vspace{-2pt}
\caption{High-level planning and reasoning evaluation results on the MetaAD dataset. Except for AlphaDrive, which utilizes our proposed training strategy, all other models are trained based on SFT. ${\text{\textdagger}}$ denotes fine-tuned on the MetaAD dataset.}
\label{tab:main}
% \vspace{-1em}
\end{table*}

\begin{table*}[]
\begin{center}
\centering
\resizebox{0.98\textwidth}{!}{
\begin{tabular}{c|ccccc|c|ccc|cccc}
\toprule
\multirow{2}{*}{ID} & Base & Plan. & Action & Plan. & Plan. & \multirow{2}{*}{Acc. $(\%)$} & \multicolumn{3}{c|}{Path $($F1$)$ $\uparrow$} & \multicolumn{4}{c}{Speed $($F1$)$ $\uparrow$} \\
& Acc. & Acc. & Weighted & Diversity & Format & & straight & left & right & keep & acc. & dec. & stop  \\
\midrule
1 & \cmark &  & & & & 42.36 & 69.40 & 64.42 & 59.02 & 62.18 & 23.72 & 47.48 & 62.70 \\
2 & \cmark & & & & \cmark & 55.71 & 83.19 & 77.34 & 71.65 & 67.37 & 34.07 & 59.87 & 76.56 \\
3 & & \cmark & & & \cmark & 67.91 & 91.95 & 82.65 & 88.01 & 77.74 & 49.79 & 61.38 & 85.75 \\
4 & & \cmark & \cmark &  & \cmark & \underline{72.20} & \underline{95.93} & \underline{85.39} & \underline{88.80} & 82.54 & \underline{52.64} & \underline{67.60} & \textbf{86.76} \\
5 & & \cmark &  & \cmark & \cmark & 69.38 & 92.10 & 80.48 & 85.59 & \underline{84.53} & 49.40 & 64.07 & 83.34 \\
6 & & \cmark & \cmark & \cmark & \cmark & \textbf{77.12} & \textbf{96.62} & \textbf{89.83} & \textbf{93.25} & \textbf{86.80} & \textbf{56.33} & \textbf{71.40} & \underline{86.63} \\
\bottomrule
\end{tabular}
}
\end{center}
\vspace{-2mm}
\caption{Ablations on the effectiveness of our proposed planning GRPO rewards.}
\label{tab:reward}
\end{table*}

\boldparagraph{Planning Diversity Reward.}
Since planning is inherently multimodal, during GRPO-based RL training, the model generates multiple solutions for group relative optimization. In the later stages of training, we observe that the model's outputs tend to converge to the same solution. Our goal is to encourage the model to generate a variety of feasible solutions, rather than merely aligning with the ground truth actions in the training data. To achieve this, we propose the Planning Diversity Reward. When the model's outputs differ, we assign a higher reward; otherwise, we reduce the reward.

\boldparagraph{Planning Format Reward.}
The last reward is used to regularize the output, making it easier to extract both the reasoning process and the final answer. This approach is inspired by R1. The reasoning process is encapsulated within the \texttt{<think></think>} tags, while the planning result is enclosed within the \texttt{<answer></answer>} tags. If the final output does not conform to this format, the format reward will be set to 0.

The Planning Accuracy Reward, the Action-Weighted Reward, and the Planning Diversity Reward are multiplied to compute the Planning Quality Reward. We calculate the Planning Quality Reward separately for speed planning and direction planning. Finally, the Planning Quality Reward and the Planning Format Reward are used to calculate the GRPO loss and update the model parameters. For details about Planning Reward Modeling, please refer to Alg.~\ref{algo:reward}.

\subsection{Reasoning: Distillation from Large Models}
Unlike fields such as mathematics or science, which have abundant high-quality reasoning data available for training, the planning process in autonomous driving is difficult to record, and the cost of manual annotation is high. As a result, there is currently no large-scale, readily available planning reasoning dataset. We initially attempt to incorporate reasoning steps directly into the RL training process, but the final results are suboptimal, mainly due to the following shortcomings: (1) insufficient perception of key elements, such as traffic lights; (2) disorganized reasoning process with weak causal relationships; (3) reasoning outputs that are overly lengthy and ineffective.

Therefore, we adopt a more capable cloud-based large model, such as GPT-4o, to generate high-quality planning reasoning data from a small set of driving clips. Specifically, we provide the model with prompts that include the real driving actions in a given scenario, along with the vehicle's current state and navigation information, prompting the model to generate a concise decision-making process. We find that the quality of the generated reasoning process is pretty good. After conducting a manual quality check and filtering out samples with obvious errors, we obtain a batch of high-quality planning reasoning data. Subsequently, our model can improve its planning reasoning ability through knowledge distillation based on this data.

% \vspace{-1mm}
\subsection{Training: SFT Warm-Up, RL Exploration}
RL relies on sparse reward signals, whereas SFT is based on dense supervision, making it more suitable for knowledge distillation. Additionally, we find that relying solely on RL can lead to instability in the early stages of training. Therefore, we use a small amount of data for a warm-up phase based on SFT, followed by RL training with the full dataset. We discover that this approach improves stability in the early stages of training and enhances the model's planning reasoning performance, ultimately leading to better overall planning capabilities.

\begin{table*}
\small
\centering
% \resizebox{0.99\textwidth}{!}{
\setlength\tabcolsep{5pt}
\begin{tabular}{c|c|c|ccc|cccc|ccc}
\toprule
With & Train. & \multirow{2}{*}{Acc. $(\%)$} & \multicolumn{3}{c|}{Path $($F1$)$ $\uparrow$} & \multicolumn{4}{c|}{Speed $($F1$)$ $\uparrow$} & \multirow{2}{*}{BLEU-4} & \multirow{2}{*}{CIDEr} & \multirow{2}{*}{METEOR}\\
Reason. & Strategy & & straight & left & right & keep & acc. & dec. & stop & & \\
\midrule
\xmark & SFT & 56.97 & 77.76 & 63.69 & 65.07 & 76.22 & 37.11 & 51.99 & 75.72 & - & - & - \\
\xmark & RL & 62.16 & 82.32 & 72.39 & 71.24 & 75.03 & 41.13 & 61.08 & 79.15 & - & - & - \\
\xmark & SFT+RL & 70.73 & 88.04 & 75.75 & 78.79 & 78.60 & 45.00 & \underline{65.92} & 83.52 & - & - & - \\
\midrule
\cmark & SFT & 65.40 & 92.52 & 71.28 & 68.65 & 81.91 & 36.48 & 59.31 & 71.55 & \underline{37.21} & \underline{34.30} & \underline{47.54} \\
\cmark & RL & \underline{72.41} & \underline{93.16} & \underline{84.24} & \underline{89.32} & \textbf{87.58} & \underline{51.19} & 64.70 & \underline{84.07} & 25.14 & 24.58 & 38.10 \\
\cmark & SFT+RL & \textbf{77.12} & \textbf{96.62} & \textbf{89.83} & \textbf{93.25} & \underline{86.80} & \textbf{56.33} & \textbf{71.40} & \textbf{86.63} & \textbf{43.54} & \textbf{38.97} & \textbf{55.23} \\
\bottomrule
\end{tabular}
% }
% \vspace{-2pt}
\caption{Ablations on different reasoning training strategies.}
\label{tab:reason}
% \vspace{-1em}
\end{table*}

\begin{table*}
\small
\centering
% \resizebox{0.99\textwidth}{!}{
\setlength\tabcolsep{5pt}
\begin{tabular}{c|c|c|ccc|cccc|ccc}
\toprule
Train. & Train. & \multirow{2}{*}{Acc. $(\%)$} & \multicolumn{3}{c|}{Path $($F1$)$ $\uparrow$} & \multicolumn{4}{c|}{Speed $($F1$)$ $\uparrow$} & \multirow{2}{*}{BLEU-4} & \multirow{2}{*}{CIDEr} & \multirow{2}{*}{METEOR}\\
Data & Strategy & & straight & left & right & keep & acc. & dec. & stop & & \\
\midrule
20k & SFT & 41.12 & 56.15 & 36.72 & 35.59 & 40.63 & 17.14 & 16.74 & 19.19 & 27.18 & 15.42 & 31.17 \\
20k & RL & 45.46 & 69.28 & 59.42 & 51.91 & 56.93 & 30.82 & 37.71 & 30.94 & 20.33 & 11.01 & 23.09 \\
20k & SFT+RL & 55.64 & 68.25 & 64.06 & 56.87 & 58.61 & 45.19 & 53.68 & 44.09 & 32.84 & 17.02 & 35.93 \\
\midrule
50k & SFT & 53.02 & 73.74 & 62.45 & 65.43 & 70.07 & 33.83 & 38.94 & 53.96 & 34.48 & 26.83 & 42.85 \\
50k & RL & 59.33 & 77.69 & 68.55 & 73.82 & 77.05 & 40.72 & 45.20 & 57.06 & 22.37 & 16.81 & 25.81 \\
50k & SFT+RL & 70.83 & 82.30 & 78.05 & 82.17 & \underline{84.80} & 47.27 & 58.29 & 64.67 & 32.30 & 30.38 & 46.38 \\
\midrule
110k & SFT & 65.40 & 82.52 & 71.28 & 68.65 & 81.91 & 36.48 & 59.31 & 71.55 & \underline{37.21} & \underline{34.30} & \underline{49.54} \\
110k & RL & \underline{72.41} & \underline{93.16} & \underline{84.24} & \underline{89.32} & 82.58 & \underline{51.19} & \underline{64.70} & \underline{82.02} & 25.14 & 24.58 & 38.10 \\
110k & SFT+RL & \textbf{77.12} & \textbf{96.62} & \textbf{89.83} & \textbf{93.25} & \textbf{86.80} & \textbf{56.33} & \textbf{71.40} & \textbf{86.63} & \textbf{43.54} & \textbf{38.97} & \textbf{55.23} \\
\bottomrule
\end{tabular}
% }
% \vspace{-2pt}
\caption{Ablations on the amount of training data.}
\label{tab:data}
% \vspace{-1em}
\end{table*}

\section{Experiments}

\subsection{Experimental Settings}
\boldparagraph{Dataset.}
We adopt MetaAD, a large-scale real-world driving dataset, as our training and evaluation benchmark. This dataset consists of a total of 120k driving clips, each lasting three seconds. MetaAD is a high-quality dataset specifically designed for planning, supporting multi-sensor data and perception annotations. Furthermore, it maintains a well-balanced distribution across various driving environments and planning actions. The dataset is divided into 110k clips for training and 10k clips for validation. As for reasoning, we sample 30k data from the training dataset to generate the planning reasoning process. All reported results are obtained by training on the training set and evaluating on the validation set.

\boldparagraph{Training.} 
We use Qwen2VL-2B~\cite{wang2024qwen2vl} as the base model. Qwen2VL is currently one of the best-performing open-source models, and it offers a smaller 2B version that better meets the latency requirements for autonomous driving. Additionally, Qwen2VL provides better support for RL. The model's inputs include a front-view image and a planning prompt, which contains the vehicle’s current speed and navigation information. The navigation data, consistent with real-world driving, is obtained from sparse navigation points via AMap (similar to Google Maps) and is converted into text form for inclusion in the prompt, such as \textquotedblleft Go straight for 100m, then turn right\textquotedblright. Training is conducted using 16 NVIDIA A800 GPUs.

\boldparagraph{Evaluation.} 
The evaluation metrics consist of two aspects. First, the accuracy of meta-action planning is measured by calculating the F1-Score for all categories of lateral and longitudinal meta-actions, followed by the overall planning accuracy. Additionally, for planning reasoning, we compute the similarity between the generated planning reasoning process and the annotated reasoning process in the dataset using BLEU-4~\cite{papineni2002bleu}, CIDEr~\cite{vedantam2015cider}, and METEOR~\cite{banerjee2005meteor} scores.

\begin{figure*}[ht]
\centering
% \vspace{1mm}
\includegraphics[width=0.98\textwidth]{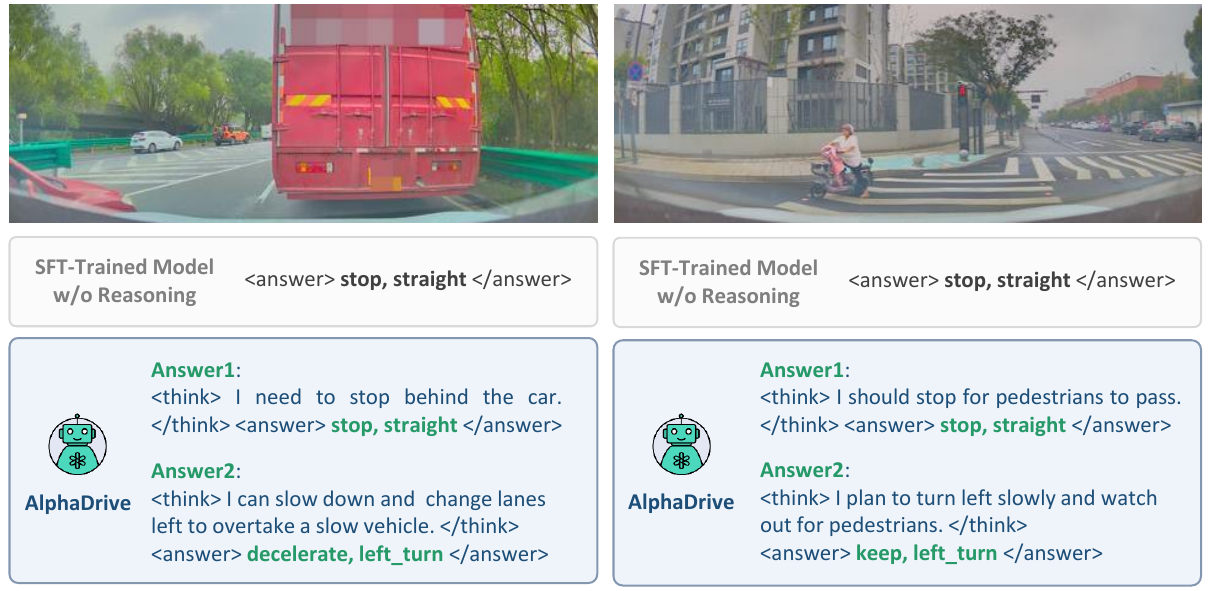} % Reduce the figure size so that it is slightly narrower than the column.
\caption{Qualitative results of AlphaDirve. After RL training, AlphaDrive exhibits some emergent multimodal planning capabilities, which has great potential for improving driving safety and efficiency.}
\label{fig:vis}
\end{figure*}

\subsection{Main Results}
Tab.~\ref{tab:main} presents the performance of AlphaDrive in high-level planning. The first four rows show the results obtained by directly evaluating the corresponding pretrained models. It can be observed that, while these models demonstrate stronger general capabilities, their performance in planning is suboptimal, highlighting the need for further training with driving data. The subsequent five rows display the results of models fine-tuned on the MetaAD dataset. As shown, AlphaDrive significantly outperforms the other models. Compared to Qwen2VL-7B, the second-best performing model after AlphaDrive, the planning accuracy significantly improves by 25.5\%. There is a noticeable enhancement in key decisions such as steering and acceleration/deceleration. Additionally, the quality of planning reasoning is the best among all models, demonstrating the effectiveness of our proposed two-stage RL training and reasoning strategies.

\subsection{Ablation Study}
\boldparagraph{Planning Rewards.}
In Tab.~\ref{tab:reward}, we validate the effectiveness of the four proposed GRPO planning rewards. The Base Accuracy reward directly determines the reward based on whether the response exactly matches the ground truth, a common approach in general domains. As shown, the model using the Base Accuracy reward lags significantly behind across all metrics (ID 1). The combination with the Planning Format Reward yields a slight improvement. (ID 2). A significant improvement is seen with the adoption of our proposed Planning Accuracy Reward (ID 3). Further enhancement in acceleration/deceleration decisions is achieved by incorporating the Action-Weighted Reward (ID 4). Finally, by combining the Planning Diversity Reward, the best planning performance is achieved (ID 5-6).

\boldparagraph{Reasoning Training Strategies.}
The ablation study of the reasoning training strategies is shown in Tab.~\ref{tab:reason}. As observed, introducing planning reasoning under different training strategies effectively enhances model performance. Notably, the improvement is especially significant for complex actions such as acceleration and deceleration, demonstrating that reasoning can greatly enhance decision-making in complex scenarios. Furthermore, the model trained exclusively with RL performs worse in reasoning compared to the model trained with SFT. We attribute this to the limited parameter size of smaller models, which results in insufficient perception and reasoning capabilities. Therefore, incorporating SFT as a warm-up phase and using knowledge distillation to learn the reasoning process from a larger model can effectively address this issue. By combining SFT and RL, the model achieves the best planning reasoning capabilities.

\boldparagraph{Amount of Training Data.}
Tab.~\ref{tab:data} shows the impact of training data size on different training strategies. As observed, when the training data size decreases, SFT is more affected. With only 20k training samples, the model trained with RL reaches a planning accuracy of 46.08\%, which is significantly higher than that of the SFT-trained model. When using nearly half of the data, with 50k samples, AlphaDrive already achieves a planning accuracy of 70.83\%, demonstrating the efficiency of our training strategy.

\subsection{Emergence of Multimodal Planning Capability}
Fig.~\ref{fig:vis} illustrates the multimodal planning capability of AlphaDrive after RL training. In complex scenarios, it can effectively generate multiple feasible solutions, whereas the SFT-trained model can only produce a single planning decision. AlphaDrive can be integrated with a downstream action model to dynamically select the optimal solution from multiple options.

% \vspace{-3mm}
\section{Conclusions and Limitations}

In this work, we propose AlphaDrive, a VLM for high-level planning in autonomous driving. Compared to previous models that solely employed the SFT, we explore the integration of advanced RL and reasoning in planning. Specifically, AlphaDrive introduces a planning-oriented RL strategy based on GRPO and further designs a two-stage planning reasoning training paradigm. To the best of our knowledge, AlphaDrive is the first to introduce the RL and reasoning to autonomous driving planning, significantly boosting both performance and training efficiency.

Currently, due to a lack of rich data annotation, AlphaDrive is still unable to output more complex driving behaviors such as lane changes or nudges. Additionally, the current planning reasoning data come from pseudo-labels generated by large models based on ground-truth driving actions, which still suffer from inaccurate perception and a failure to capture key factors. Therefore, further systematic validation is required to improve data quality and verify the performance upper bound of \thename{}.

\section*{Acknowledgments}
\setlength{\emergencystretch}{0.8em}
We sincerely thank Hao Gao, Tianheng Cheng, Bencheng Liao, Haoyi Jiang, and Dongli Hu for their valuable feedback on the draft.

{
    \small
    \bibliographystyle{ieeenat_fullname}
    \bibliography{main}
}

\end{document}